# Hardware Trojan Detection Using Unsupervised Deep Learning on Quantum Diamond Microscope Magnetic Field Images

MAITREYI ASHOK, Massachusetts Institute of Technology, USA
MATTHEW J. TURNER, University of Maryland, USA
RONALD L WALSWORTH, University of Maryland, USA
EDLYN V. LEVINE, Harvard University, USA, The MITRE Corporation, USA, and University of Maryland, USA
ANANTHA P. CHANDRAKASAN, Massachusetts Institute of Technology, USA

This paper presents a method for hardware trojan detection in integrated circuits. Unsupervised deep learning is used to classify wide field-of-view (4x4 mm$^2$), high spatial resolution magnetic field images taken using a Quantum Diamond Microscope (QDM). QDM magnetic imaging is enhanced using quantum control techniques and improved diamond material to increase magnetic field sensitivity by a factor of 4 and measurement speed by a factor of 16 over previous demonstrations. These upgrades facilitate the first demonstration of QDM magnetic field measurement for hardware trojan detection. Unsupervised convolutional neural networks and clustering are used to infer trojan presence from unlabeled data sets of 600x600 pixel magnetic field images without human bias. This analysis is shown to be more accurate than principal component analysis for distinguishing between field programmable gate arrays configured with trojan free and trojan inserted logic. This framework is tested on a set of scalable trojans that we developed and measured with the QDM. Scalable and TrustHub trojans are detectable down to a minimum trojan trigger size of 0.5% of the total logic. The trojan detection framework can be used for golden-chip free detection, since knowledge of the chips' identities is only used to evaluate detection accuracy.

CCS Concepts: • **Hardware** → *Reconfigurable logic and FPGAs*; *Hardware reliability screening*; **Quantum technologies**; • **Security and privacy** → **Malicious design modifications**.

Additional Key Words and Phrases: hardware trojan detection, quantum diamond microscope, convolutional neural networks

## 1 INTRODUCTION

Security is a necessary parameter of microelectronic technology development due to the harmful effects an attacker can have on a system's operation. One possible attack is hardware trojan insertion, consisting of adding malicious circuitry during a step in the supply chain cycle to provide unintended functionality, such as information leakage, denial of service, or behavior modification [5, 7]. Hardware trojans at the integrated circuit (IC) level cannot be inserted once a chip is fabricated and packaged. However, if a trojan is detected later in the supply chain, the entire batch of chips needs to be discarded. Firmware patches are possible in some cases, but this is not a guaranteed fix since it is possible a backdoor has already been established [5, 34].

With the pervasiveness of ICs in almost every device and system, hardware trojans have emerged as an increasingly likely and dangerous vulnerability. This threat is compounded by most equipment requiring at least some commercial circuits, the design of many circuit blocks being outsourced, and the fabrication and packaging of chips by third parties.

Reliable and efficient detection of hardware trojans is consequently imperative. Many methods leverage nondestructive functional testing [7, 8]. However, trojans will remain undetected if functional testing inputs and conditions do not match trigger conditions. Other destructive methods successively scan layers of the circuit and match to the layout to detect any deviation. However, these cannot be applied to every fabricated chip, and a clever attacker may only insert the trojan in a subset of a batch.

In this work, we introduce a new, non-destructive method of detecting trojans, shown in the system block diagram in Figure 1.

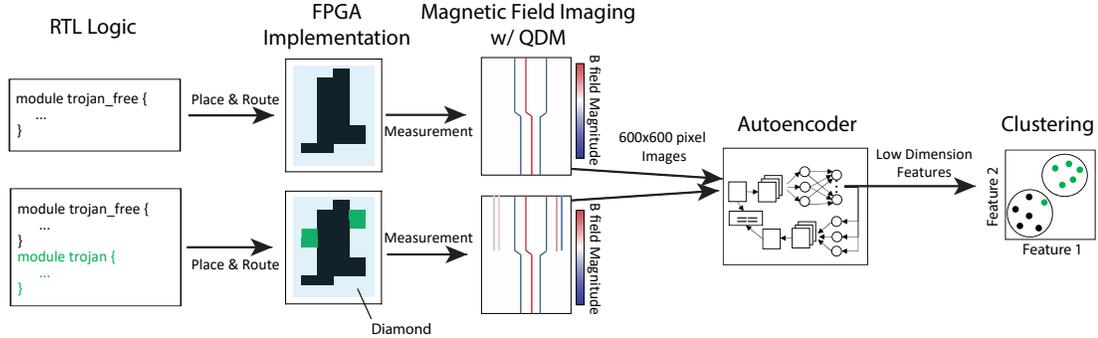

Fig. 1. Scalable trojan free circuits and trojans are designed at the register transfer level (RTL). They are placed and routed for the field programmable gate array (FPGA) implementation. The Quantum Diamond Microscope (QDM) is used for electromagnetic (EM) side channel measurements of the FPGA. The convolutional neural network (CNN) autoencoder analyzes the measurements and creates a small set of latent variables, which are then clustered for trojan detection.

### 1.1 Threat Model

Hardware trojans can be inserted during the design, fabrication, or packaging stages. Trojans added during the design stage can be discovered through careful analysis of the netlist or GDS data [11]. However, trojans inserted during the fabrication stage are much more difficult to detect. Every chip produced by the foundry is untrusted, and there is no golden version to compare against. Furthermore, if the financial benefit of a trojan attack outweighs the cost of having multiple masks (e.g. the trojan will leak bank account information or interfere with military operations), a foundry may only insert the trojan into a subset of a batch of chips. The motivation for this is provided in Figure 2. In this case, it is not enough to destructively analyze a few chips and conclude that the entire batch is trojan-free.

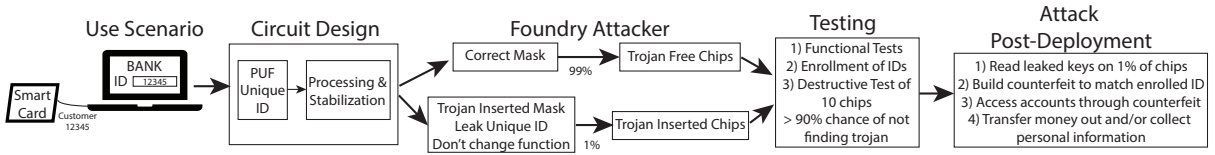

Fig. 2. Example scenario of trojan insertion into bank login smart cards by a bad actor at the foundry, where a batch of at least 100 chips is being produced. By inserting a trojan only into a small subset of chips with a separate mask, the attacker has a high probability of evading detection through standard tests. Even with a small percentage of counterfeit-able chips, the potential monetary gain for the attacker can be significant, especially with a large bank.

The threat model for this work is an attacker at the foundry inserting trojans during the fabrication of a submitted chip design, similar to the cases proposed in [1, 21]. [21] shows the possibility of adding harmful trojans directly to IC mask layout files with very little time or knowledge of the original circuit. The foundry does not need to keep track of which chips are HT-inserted; rather, an attack could later automatedly be manifested in all of the HT-inserted chips, without any manual trigger necessary.

In this threat model, trojans added during the fabrication stage can only cause minimal changes in placement and routing. In this case, major changes as with those done with CAD tool algorithms cannot be done, since the attacker risks breaking the normal operation of the circuit with too many changes. Thus, most useful trojans will only involve the addition of a small footprint trigger and payload, such as a comparator or a few registers, where the payload becomes active when a rare trigger condition is met. We focus on the detection of these trojans in digital integrated circuits, with a small trigger circuit insertion. To allow for quick turnaround time between trojan design and detection, we use trojans defined at the register transfer level (RTL) of a field programmable gate array (FPGA) as a proxy.

We do not consider trojans with no addition or removal of logic, and only minor changes in routing, since these do not have the potential for complex attacks that would not be evident in functional tests. Analog circuit-based trojans, doping level trojans [4], and similar non-digital attacks are out of the scope of this work as well.

In the final part of the threat model, we assume that once the original designers receive the dies, they are able to run any functional or side channel tests on all of the chips. Furthermore, the designers have all necessary GPU or high performance computing capabilities, and are not limited in what algorithms can be used to analyze side channel measurement data. In addition, they may perform destructive analysis for full layout comparison on a very small subset of chips.

## 1.2 Contributions

This paper improves upon existing developments in hardware trojan detection by introducing a new sensing modality to the field, the Quantum Diamond Microscope (QDM), which has a wide field-of-view (∼millimeters) and is able to measure vector magnetic fields at a much finer spatial resolution (∼microns) compared to the traditional electromagnetic (EM) field probe. In addition, the QDM provides a quantitative measurement of the circuit activity, compared to thermal imaging or other similar techniques. In summary, the main contributions of this work are

- **Demonstrating the first use of the QDM for trojan detection and improving the QDM setup (from [33]) to achieve better spatial resolution and faster measurements via improved magnetic field sensitivity.**
- **Reliably detecting small footprint hardware trojans with machine learning (ML) on high spatial resolution magnetic field measurements taken with the QDM. The detection ability with deep learning is shown to be higher than with the traditional principal component analysis (PCA).**
- **Performing automated trojan detection using unlabeled datasets with clustering algorithms.**
- **Developing a set of base trojan free circuits and scalable trojan benchmarks for better evaluation of trojan detection ability and limits.**

## 2 BACKGROUND

### 2.1 Hardware Trojan Detection

Prior research on nondestructive methods for hardware trojan detection spans functional testing, side channel testing, and use of localized detection circuitry within the die. Functional testing can find trojans that are activated with small triggers which are likely to be hit, and that have an obvious impact on the chip. Unfortunately, this testing may miss many trojans since attackers can design the trojan to evade detection through simplistic tests [7, 36].

Side channel testing allows for detection of trojans that are not activated during the test time. By measuring side channels such as power or EM radiation of a test chip and a golden circuit, trojan ICs can be detected [2, 22]. The usefulness of side channel detection is limited by the sensor's ability to detect potentially small trigger activity in the IC. We can improve detection ability of this method by increasing the sensitivity of the measurement tool [5], for example by using a QDM instead of a field probe.

Furthermore, many side channel based trojan detection methods require comparison to a golden circuit [2]. However, a golden circuit is not always available. Thus, we must develop methods of analysis that can detect the presence of abnormal chip behavior even in the absence of a golden circuit, for example with clustering algorithms.

Prior results also suggest that localizing measurements to small regions of the IC helps with trojan detection [13, 26]. These methods add circuitry such as ring oscillators or sleep transistors in the die itself. The results of adding this additional circuitry are then measured by external side channel probes. However, an attacker can modify the circuitry placed for sensitivity analysis and cancel the designer's protection. Thus, localized detection in small regions of the IC should be achieved from a measurement technique external to the die. With a standard EM field probe, the low spatial resolution from the cross section limits localization. This limitation can be overcome with more advanced magnetic field imaging with the QDM.

### 2.2 Machine Learning (ML) Analysis Methods

Previous EM-side channel based hardware trojan detection methods use a near-field EM probe stepped over an IC to record localized EM traces during circuit operation. The measurements are then compared between the golden circuit and test circuit using correlation analysis or PCA to detect trojans [2, 15, 23, 30]. Automated detection with PCA has also been explored using backscattering side channels [24] and for netlist data [11]. However, PCA cannot capture many complex or nonlinear dependencies in physical measurement data, especially as dimensionality increases with spatial resolution.

There have been some works considering more complex ML methods, including support vector machines and neural networks [16, 21, 35]. These works generally use supervised learning with the assumption that all of the training data is labeled as trojan free or trojan inserted. However, the designer who receives chips from an untrusted foundry will not be given a list of which ICs were attacked, and the specific trojans that were inserted. Thus, the designer will not have any knowledge of what labels to provide for different chips when training. For this reason, unsupervised learning with label-free training data is required, and a basic classifier algorithm cannot be used.

### 2.3 Quantum Diamond Microscope (QDM)

The QDM provides vector magnetic field imaging with high spatial resolution (~microns) over a wide field-of-view (~millimeters), and can be used for side channel measurements. The QDM uses internal quantum transitions of optically-active nitrogen vacancy (NV) centers in a diamond chip placed above an IC to measure circuit activity due to differential current flows in the IC. Zeeman splitting causes NV energy levels to differ based on the spatially-varying magnetic fields created by currents in the IC. The magnetic field distribution from the IC is determined by measuring the NV fluorescence contrast for different microwave frequencies with a camera [18], essentially providing a sensitive, high-resolution optical image of the local vector magnetic fields produced by the active IC.

The QDM simultaneously measures over a wide field-of-view of the silicon die, generating, in the current measurements, 600 × 600 pixel magnetic field images, where each pixel is approximately 6$\mu$m. This is done with a diamond of approximate size 4 mm × 4 mm × 0.5 mm that has NVs located only at the surface of the diamond chip (depth of few microns). The QDM provides a high spatial resolution magnetic image on the order of a few microns. The setup is currently optimized for monitoring static magnetic fields and the magnetic signal is averaged for 1 to 5 minutes to simultaneously obtain sufficient SNR for all ∼ $4 \times 10^5$ pixels in the field-of-view. New techniques have demonstrated the ability to perform real time magnetic imaging over a wide field-of-view at kHz frame rates [32]; these capabilities will be applied to ICs in future work. Table 1 compares the QDM sensor used in the present study to other side channel detection methods. The QDM provides value over traditional EM [2, 15] or power [6] side channel techniques through simultaneous imaging with high spatial resolution over a wide field-of-view. Furthermore, we envision the use of the QDM + machine learning as a complementary method to existing hardware trojan detection methods. In particular, one could combine measurements from multiple methods to represent various facets of the data: e.g., combined high spatial resolution and wide field-of-view with the QDM and high temporal resolution with a near field probe. We plan to explore this multi-modal approach in future work.

Table 1. Comparison between different methods of side channel detection.

|  | Power | EM Probe | QDM [33] |
| --- | --- | --- | --- |
| Spatial Resolution | None | 60 $\mu$m [12] | 5-10 $\mu$m |
| Sensitivity | Limited by number of bits in oscilloscope | Limited by number of bits in oscilloscope | Noise floor ∼ 0.1 $\mu$A |
| Time Resolution | Limited by oscilloscope sampling frequency | Limited by oscilloscope sampling frequency, Probe bandwidth 1.5 MHz up to 6 GHz [12] | Limited by averaging period for desired SNR |
| Field-of-View | N/A | Single-point Measurement | 3.7 mm × 3.7 mm |

The QDM+ML method has previously demonstrated imaging of ring oscillators in a FPGA, with small numbers of ring oscillators detected within a wide field-of-view [33]. However, the setup must be optimized for trojan detection since well-designed trojans are expected to have less switching activity and lower currents than a ring oscillator.

To the best of our knowledge, this is the first example of hardware trojan detection where the spatial structure from physical measurements is preserved during the ML process and where unsupervised deep learning and clustering are used for unbiased analysis.

## 3 EXPERIMENTAL SETUP

### 3.1 QDM Setup

The experimental setup in Figure 3 is a modified version of that in Ref. [33] (photos of the setup are in Appendix C). The QDM employs a ∼ 4 mm by 4 mm by 0.5 mm diamond substrate with a 1.7 $\mu$m thick N-doped, isotopically purified (99.995 % $^{12}C$) layer on the top surface with [$^{15}N$] = 17 ppm and [$NV$] = 2 ppm. The NV-layer is illuminated with 1 W of 532 nm excitation laser light spread out over a 5 mm by 5 mm beam spot, encompassing the entire diamond, using total internal reflection to limit exposure of the IC to green light. The green illumination is used to both initialize the NV centers into a well defined spin state ($m_s$ = 0 in the ground state) and to readout the spin state through the spin-state dependent fluorescence contrast, a process known as optically detected magnetic resonance (ODMR) [3, 31].

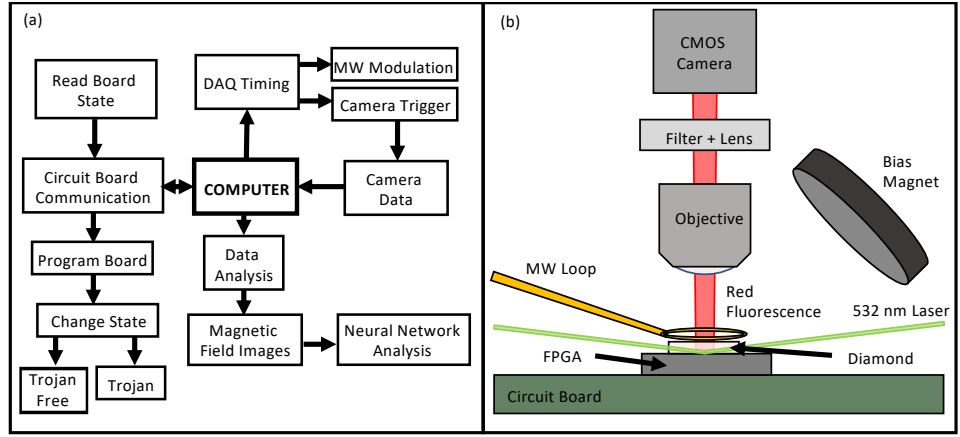

Fig. 3. (a) Flowchart of relevant communication and protocols for the QDM magnetic imaging experiments.
(b) Schematic of experimental components for the QDM experiments. The NV diamond is placed directly on an FPGA. Excitation from a 532 nm shines at a shallow angle and the resultant red fluorescence is collected onto a CMOS camera. The NVs are controlled through the application of external microwaves and a DC bias field.

A pair of permanent SmCo magnets is used to apply a static bias magnetic field of $\mathbf{B}_0 = (B_X, B_Y, B_Z) = (2.0, 1.6, 0.7)$ mT to lift the degeneracy of the four NV crystallographic axes and enable the probing of a fixed crystallographic axis through controlled microwave (MW) driving. A wound copper wire loop (5 mm in diameter) is used to apply a ∼ 10 $\mu T$ driving MW field near 2.87 GHz which resonantly drives the NV spin state from the $m_s = 0$ to the $m_s = -1$ or the $m_s = 0$ to the $m_s = +1$ spin state. In contrast to the setup in Ref. [33], a double-balanced mixer is used to generate sidebands which create two MW tones separated by 3.05 MHz to simultaneously drive the hyperfine spin transitions of the NVs [32]. This hyperfine driving technique improves the fluorescence resonance contrast and minimizes the range of MW frequencies needed to fully address a resonance feature, speeding up a given measurement sequence. More detail about the experimental components and measurement protocols can be found in Ref. [33] and [32].

The MW sweeping is synchronized with frame acquisition of the camera and state control of the external FPGA under test. The resultant NV fluorescence resonances are fit pixel by pixel to Lorentzian curves. The resonance positions of the lower transition ($m_s = 0$ to the $m_s = -1$) and upper transition ($m_s = 0$ to the $m_s = +1$) are used to extract the time-averaged magnetic field and temperature over a given measurement cycle. The improved diamond substrate and hyperfine driving technique employed in this work enhance the ODMR contrast and thus the magnetic field sensitivity of the QDM by a factor of ∼ 4, enabling an increase in measurement speed by a factor of ∼ 16 (or an increased measurement SNR for a fixed measurement duration), compared to previous QDM IC demonstrations [33].

### 3.2 FPGA Setup

A FPGA is used to model trojans that are added during the fabrication of an IC [2, 22, 23, 30]. This provides a fast turnaround time between trojan design and trojan detection tests, with the process shown in Figure 1. For trojan insertion during mask generation and fabrication, there is limited ability to move existing logic while ensuring expected basic functionality. Thus, additional trojan logic mostly consists of extra interconnect and logic within filler cells [21, 36].

The FPGA used is the Xilinx Artix-7 XC7A100T. The chip is decapsulated to minimize the distance between the NV centers and the silicon die, which improves spatial resolution of QDM magnetic images. The present work is an early stage demonstration of the applications of the QDM+ML method to hardware trojan detection. Future efforts will include extensions of our approach to intact FPGA samples. Note that prior work with the QDM+ML method has shown the ability to measure and characterize circuit activity noninvasively, with only slight reduction in accuracy compared to invasive measurements [33]. We expect that similar results will hold for applications of the QDM+ML method to hardware trojan detection.

To ensure the FPGA trojan is a close model to a hardware trojan in a fully custom IC, we use Vivado [37] to lock the placement and routing of the trojan free circuit logic. Trojan logic is subsequently added in any surrounding regions with the constraint that the original logic not move. In addition, extraneous logic in the relevant FPGA field-of-view such as the analog-to-digital converter and the sensors are turned off.

The top metal layer(s) of the FPGA contain the power distribution network. Depending on which portions of the die consume more power, the power grid traces in certain locations will carry more current than others. These spatially distributed current carrying wires create a spatially varying magnetic field distribution above the chip, which can be measured using the QDM. This provides a "heatmap" image of the magnetic fields, with the field magnitudes at different measurement points.

When a trojan is inserted into a circuit, there are additions of logic or changes in the wiring. These cause differences in the spatial current distribution, and thus in the magnetic field. This can be measured in the QDM as imaging anomalies from the magnetic field of the trojan free circuit. An example of this is shown in Figure 4.

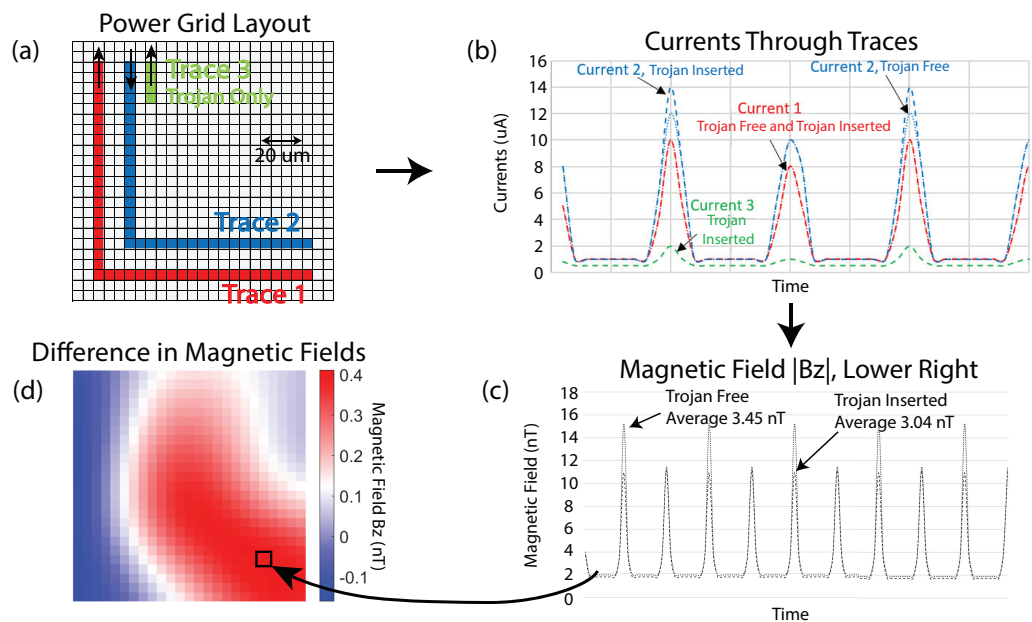

Fig. 4. An representative example of how the QDM is able to image trojans is shown for the power grid layout in (a). Trojan logic causes additional current to flow through existing power grid traces, and may lead to the additional utilization of other top metal layer traces (b). This translates to differences in the magnetic field distribution over time (c). When an average of the magnetic field at different spatial locations is taken, there are differences in the trojan free and trojan inserted configurations (d).

This same method is not expected to be limited to FPGA trojan detection, and should be applicable to the general case of application-specific integrated circuit (ASIC) trojan detection. These must also have additional logic added in filler cells and/or changes to wiring. Furthermore, ASICs have a similar structure to FPGAs, with the power grid distribution in the top metal layers. Thus, it is expected that trojans will cause changes in the top metal layer current distributions and resulting magnetic fields in ASICs (as with the FPGA). In fact, there is a likelihood of ASICs having a greater change from trojan insertion than FPGAs, due to the additional leakage current from extra logic added. We plan to apply the QDM+ML technique to ASICs in future work.

## 4 DESIGN OF SCALABLE TROJANS

### 4.1 Motivation

A set of scalable trojans is developed to evaluate QDM+ML trojan detection. Evaluation of the detection method for trojans of diminishing scale provides a quantitative measure of detection success rather than a binary result given with benchmarks such as TrustHub [28], studied later in this paper. The TrustHub benchmarks contain quite large trojans that most approaches in the literature are already able to detect. In addition, the AES switching activity and logic (from the most commonly tested TrustHub benchmark) is more complex, increasing the difficulty of interpreting physical measurements.

Additionally, scalable circuits are the building blocks of most other circuits, and are simple enough for basic estimations of the relative magnetic fields that will result. With the scalable trojans designed here, it becomes easy to relate the FPGA RTL to the physical logic, the expected relative currents, and the resulting magnetic fields. Furthermore, these simpler circuits are quite localized, and allow us to test the abilities of the QDM, which has high spatial resolution. In addition, with these scalable trojans, we can look at the effects of not just trojan size, but also trojan switching frequency, which is not generally considered in prior literature.

These scalable trojans are used to model the triggers of hardware trojans. If the triggers can be detected, then an additional payload to induce a malicious impact on the circuit will only make detection easier since there is more logic [2]. Thus, by not including payloads in the scalable trojan tests, we can model the case when the trojan is not activated.

The trojans in this work are chosen to cover a wide range of possible trojan designs. A practical trojan trigger design will require combinational or sequential logic. The comparator trojan is chosen as a general case of combinational logic and the shift register trojan is chosen as a general case of sequential logic. The counter trojan is used for a combination of both logic types. Furthermore, since the QDM+ML method is a novel tool for integrated circuits, the scalable trojans allow us to evaluate the sensitivity to a wide variety of common circuit components and logic.

### 4.2 Trojan Free Circuit and Trojan Design

*4.2.1 Trojan Free Circuits.* One major portion of the scalable trojan tests is the baseline trojan free circuit, which represents normal operation of the IC. The two types of baseline circuits used for this study are a counter and linear feedback shift register. The first trojan free circuit evaluated is a counter. The dynamic current, and thus magnetic field, are proportional to the switching activity as a first order approximation. The counter outputs have a regular switching activity that can be estimated mathematically from the bit position. Thus, if the counter output of the baseline circuit is used as input for the trojan, the expected detectability of a trojan from a magnetic field measurement can be estimated. In contrast to other base circuits, the counter allows for easy testing of the dependence of detectability on trojan switching frequency. Multiplying or dividing the trojan input signals' frequencies by factors of 2 is equivalent to using more or less significant bits of the counter. In this work, a 200 bit counter trojan free circuit is used.

The other type of trojan free circuit used is a linear feedback shift register (LFSR). The LFSR, which is used for pseudo-random number generation in some applications, has somewhat random switching activity. If the state (shift register bits) is used as input for the trojan, all the bits chosen are approximately equivalent in terms of the switching frequency. This is a good model for what inputs from a cryptographic circuit such as AES or RSA might look like. A 167 bit LFSR is used in this paper.

*4.2.2 Comparator Trojan.* One of the scalable trojans used is a comparator, shown in Figure 5. Comparators can be used in trojan triggers such as those that check for a set of bits of the key in a cryptographic circuit to match a certain value before they start to leak information through a covert channel [5, 7]. The comparator uses purely combinational logic, and is constructed with only LUTs in the FPGA. The comparator trojan building block used in this work has inputs from four different bits of the base code count, and checks if all the bits equal 1.

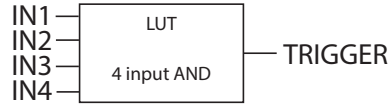

Fig. 5. Block diagram of comparator trojan.

*4.2.3 Shift Register Trojan.* Another trojan type that is used in this study is the shift register, shown in Figure 6. The shift register portion of the trojan models sequential logic, which is constructed with successive registers in the FPGA. In addition, some combinational logic is used to accumulate the shift register output similarly to the comparator trojan. A shift register can be used in trojan triggers that check for a time sequence of values of a state bit to match a certain value before it starts leaking information [7]. This work uses a four bit shift register, with a one bit input from the base code counter. This trojan building block outputs a 1 if the input bit is 1 over four consecutive clocks.

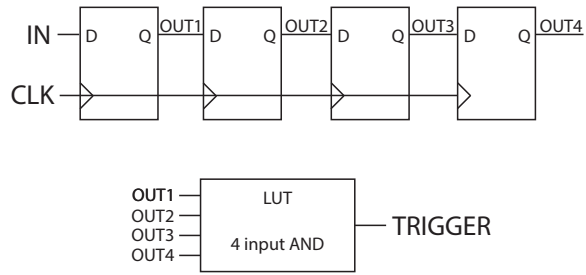

Fig. 6. Block diagram of shift register trojan.

*4.2.4 Counter Trojan.* The final trojan type used for the scalable tests is a counter, shown in Figure 7. This models a trojan that uses sequential and combinational logic for the counting operation, as well as combinational logic to accumulate values similarly to the comparator trojan. A possible scenario for a trojan with counter logic is where a count is kept of the number of times a bit of the state is set [5, 7]. Once the count reaches a certain value, information starts leaking from the chip. This type of logic can also be used for a trigger that waits for a certain amount of time (counting on the clock signal) before enabling the malicious operation. This work uses a four bit counter building block, which counts up each time the input bit from the base code state is 1. This trojan building block outputs a 1 when the count is all 1s, and then resets to a zero count on the next clock.

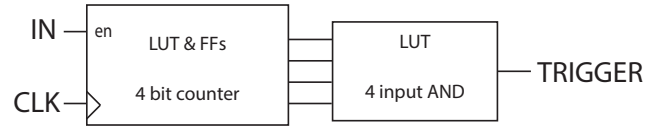

Fig. 7. Block diagram of counter trojan.

### 4.3 Construction of Tests

The scalable trojan test logic is created by constructing FPGA bitfiles with a base circuit of a size and type as well as a trojan, as seen in Figure 8. These trojans consist of one or more of the trojan building blocks described in the previous section. Each of the building blocks has its inputs connected to different state bits of the base circuit. For example, an 8 bit counter trojan has eight individual 4 bit counters, each counting the number of ones in a different bit of the base circuit count. The different trojans have different types and ratios of LUT and Register FPGA components, as well as different switching activity, which affect their magnetic field magnitudes.

As mentioned in the Experimental Setup above, placement and routing is locked between the trojan free and trojan inserted FPGA Tests. The base code and trojan for the tests are localized within a constrained region of the FPGA, or pblock, to simulate the shorter distance an attacker might want between a trojan and the original logic.

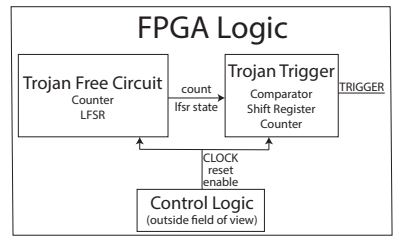

Fig. 8. Block diagram of trojan free circuit and trojan logic.

## 5 ANALYSIS METHODS

### 5.1 Principal Component Analysis (PCA)

Principal component analysis (PCA) of QDM data is evaluated here in comparison to convolutional neural networks (CNNs) to distinguish between trojan free and trojan inserted side channel measurements [14, 25]. PCA uses a linear map for dimensionality reduction, which captures the maximum variation between the training samples. In this case, the first four components are chosen, since this captures at least 50% of the explained variance during fitting. Additionally, inspection of the principal components shows that power-grid type information is only contained in the first few principal components, and this is also consistent with prior work using PCA for analysis of QDM data from IC imaging [33].

PCA reduces the dimensionality of the image data, and clustering is then applied as described in the next section to automatically decide whether the test chip contains a trojan.

### 5.2 Clustering

Density based clustering using SciKit Learn's `DBSCAN` function [25] is used for unbiased determination of whether each data point is trojan free or trojan inserted once dimensionality reduction is completed. Compared to other methods such as K-means clustering, density based clustering allows for an arbitrary number of clusters, so that the algorithm can be used without knowing whether there will only be one cluster (chips are of the same type) or multiple (two or more chip types exist). The eps value, or maximum distance between data points in the same cluster, is chosen based on the algorithm in Refs [19, 27]. This method automatedly calculates the maximum distance by finding the distances to the two nearest neighbors of each data point, and finding the point of maximum curvature, or the knee, when plotting these distances.

### 5.3 CNN Autoencoder

A convolutional neural network (CNN) autoencoder [9] is used to improve detection accuracy over the PCA method. The CNN is a deep learning technique generally used for image classification tasks that preserves spatial relationships in the data by convolving a filter with an image in each layer.

An autoencoder is used for unsupervised learning since only unlabeled data is present and it is not known in advance whether each chip is trojan inserted or not. The autoencoder takes the measured images as inputs and performs training to maximize accuracy of image reconstruction. The autoencoder convolution, pooling, and fully connected layers reduce the data to a small set of latent variables before attempting to reconstruct the higher dimensional original images. Thus, any major differences in the images must be represented in the lower dimensional data to be able to re-create the original image. By applying clustering to the low dimensional data as was done for PCA-based analysis, trojan free and trojan inserted data can be distinguished.

CNNs have been shown to outperform other techniques in the ML image analysis space [17]. CNNs have less dependence on feature engineering than other ML techniques and can generally lead to smaller networks that have comparable performance to fully connected neural networks. Furthermore, the convolutional layers preserve the 2-D structure of the images and the spatial relationships between magnetic fields from different wires [20]. In contrast, PCA flattens the image data to a 1-D list where the information on relative positions is ignored. The use of a Rectified Linear Unit (ReLU) activation function between layers in CNNs introduces an element of nonlinearity that can better model complex relationships in the data compared to PCA. In addition, some simpler techniques used for classification based learning such as support vector machines cannot be used for trojan detection since they require knowledge of trojan inserted chips for training data.

Compared to previous work, we expect that the CNNs will be particularly useful for detection with QDM magnetic field measurements, which have a high degree of spatial information compared to other EM side channel techniques. While it is expected that these qualities of CNN autoencoders will likely lead to higher accuracy compared to PCA, there are other tradeoffs for training and inference time as well as computational requirements that might necessitate PCA. In these cases, a lower detection accuracy will result; we quantify this difference in Sections 6 and 7. Furthermore, in cases of very simple datasets where PCA itself is able to perform with high accuracy, using CNNs may not be worth the additional computational cost.

Figure 9 shows the process used for training. After autoencoder training is complete, clustering can be applied to the low dimensional data for the test measurements to distinguish between trojan free and trojan inserted data.

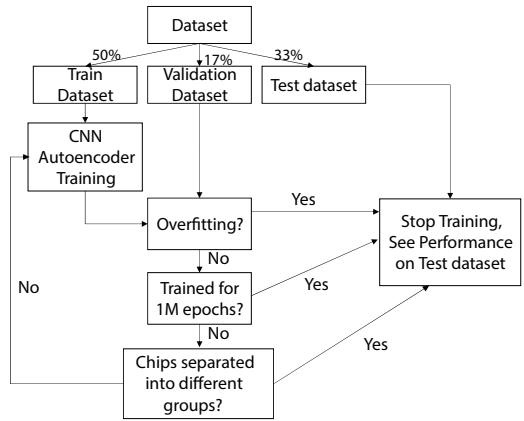

Fig. 9. Process used for training and testing the CNN autoencoder. At intervals of 100-200K epochs, training is stopped if clustering the latent variables transformed from training and validation data is split into two or more clusters. This is because the model has detected a trojan inserted in the test chip, and further training will only find finer-grain distinctions. Once the model training is done, the performance is evaluated on the test dataset.

### 5.4 Golden-Chip Free Methodology

The detection method in this work is a golden-chip free method, as in [24]. The unsupervised machine learning and clustering methods do not require any labeled training data. When a golden circuit is not available to compare against, a large number of chips will be measured and the data will be analyzed with the autoencoder and clustering framework. If there are $n$ resulting clusters, $n$ chips (one from each cluster) will be selected and reverse engineered to compare against the layout with methods such as [10]. If a trojan is found in a chip, we can assume that all the chips that were part of the same cluster have a trojan inserted. On the other hand, if a trojan is not found in a chip, then all of the chips that were part of the same cluster are similarly trojan free.

If a golden circuit is available, unsupervised or partially supervised machine learning must still be used. While labeled training data is available for the trojan-free circuit in this case, it would not be present for any trojan-inserted circuit - it is impossible to predict what specific trojan an attacker may insert. However, the destructive analysis of selected chips from each cluster is not necessary, and it is enough to check whether all the chips lie in the same cluster as the golden circuit.

In the following results, we compare QDM magnetic field measurements of the trojan inserted circuits with the corresponding original circuits. The detection framework does not use any knowledge of whether each measurement is of a trojan inserted or trojan free chip, and is thus a golden-circuit free methodology. This information is only used when evaluating the accuracy of our trojan detection method.

## 6 RESULTS

### 6.1 QDM Images

Quantum diamond microscope (QDM) magnetic imaging is used to measure magnetic side channel information from the different trojan configurations. Most of the features in the imaged magnetic fields correspond to varying currents in the top metal layer power distribution network, which lies directly below the diamond (see Figure 3). Each test configuration has 80 measurements taken to allow for analysis using ML methods. Base measurements of the chip with no switching logic are also taken interleaved with these measurements to account for environmental perturbations. These 20 measurements are subtracted from the test configuration measurements to only study RTL-dependent variations. For example, Figure 10 shows the magnetic field measurements for two trojan free chips and two counter trojan inserted test chips. It can be seen that the magnetic field measurements for the two trojan free chips are quite similar. In contrast, for the trojan inserted chips, there is additional current in the power grid, which results in larger magnetic fields and substantially different field patterns. These results can be seen effectively in the images showing the difference between the magnetic field measurements of the trojan test chip and trojan free chip. Furthermore, the difference is greater for the larger counter trojan, as expected.

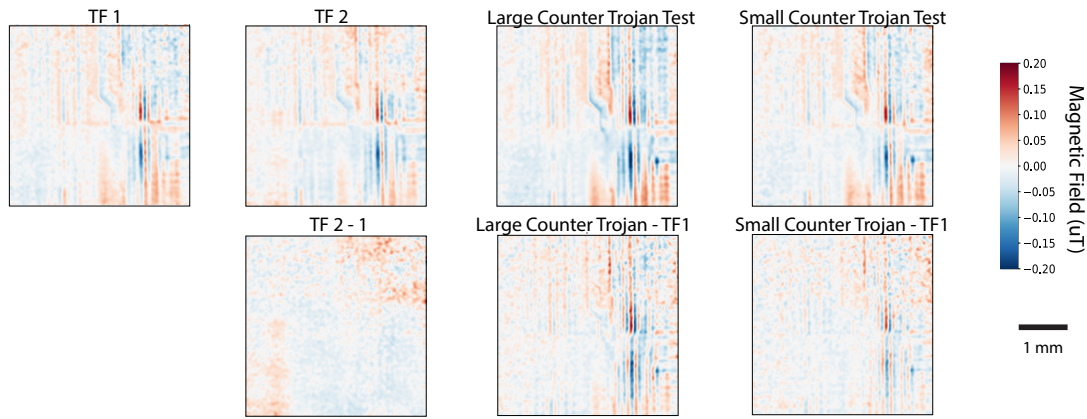

Fig. 10. QDM magnetic field images for example datasets analyzed in later sections. The images show the average of 80 measurements, and a high pass filter is applied. The bottom row shows the difference between QDM images for the test chip and one trojan free (TF) circuit.

### 6.2 Usage of Clustering Method

In the practical detection scenario, the key information would be the cluster maps for each set of data, which are indicative of how many chips there might be, and allow for more careful further analysis of an example chip from each cluster. An example of clustering with multiple chip configurations is shown in Figure 11. The algorithm, described in Section 5.2, separates the data into three clusters. Here, we choose the distance between clusters and variance for each cluster as a metric of evaluating how many different chip types might be present. We can see that the minimum distance between distinct clusters is more than three times as much as the largest cluster's standard deviation. This suggests that there are three types of chips, so we would perform destructive analysis on one chip from each of the three clusters, and compare against the layout. These results match with the original configurations, where the blue cluster consists of data from two trojan free configurations, and the green and pink clusters correspond to the small and large counter trojan, respectively, shown in Figure 10.

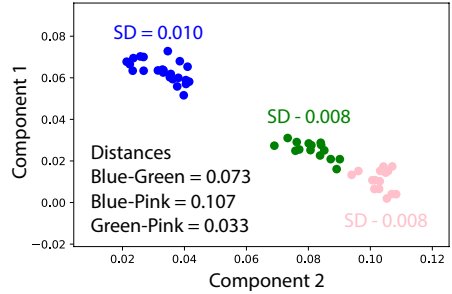

Fig. 11. Example of practical clustering with four different chips measured.

In Section 6.3 and future sections, we use a method of combining data into 2 clusters to quantify the false positives and negatives, which provides a simple figure of merit that can be compared between dimensionality reduction algorithms and different trojans.

### 6.3 Accuracy Calculation

The knowledge of whether each test chip is trojan free or trojan inserted is only used to calculate the accuracy of our detection framework. The clustering method described in Section 5.2 splits all of the measurement data points into one or more groups (clusters). These clusters are combined to only have one or two different consolidated groups, since we only compare measurements from two chips at a time in this work. If the majority of data points in a certain cluster are from chip 1, then the entire cluster is assigned to chip 1. The same is done for chip 2 clusters. Based on this, all the data points are in either one or two groups. If we only have one group, we assign that group to be trojan free.

Based on this consolidated grouping, we can calculate the trojan detection accuracy. If the test chip is actually trojan free, the number of false positives is the number of measurements that belong to a cluster only containing one of the chips. If the test chip is trojan free, false negatives do not apply.

If the test chip is actually trojan inserted, the number of false positives is the number of trojan free chip measurements in a group assigned to the test chip. Similarly, the number of false negatives is the number of test chip measurements in a group assigned to the trojan free chip.

A schematic example of the cluster consolidation and false positive/negative calculation is shown in Figure 12.

### 6.4 Preprocessing

We average successive measurements of each test configuration to improve the SNR of the magnetic field data. Figure 13 shows that averaging every two successive measurements yields the best results in terms of accuracy, number of data points categorized as noise during clustering, and the total data set size for machine learning. With the selected level of averaging of every two successive measurements, there are 80 images in the data set: 40 for the test chip (either trojan free or trojan inserted) and 40 for the trojan free circuit. A high pass spatial filter is also used to filter out RTL independent activity of low spatial variation, and to increase the contribution of the magnetic fields from power traces that have sharp spatial changes. Since the scalable trojans are smaller and more localized to one region of the FPGA, only the bottom 58% of the field-of-view (348 × 600 pixels) is used in the dataset.

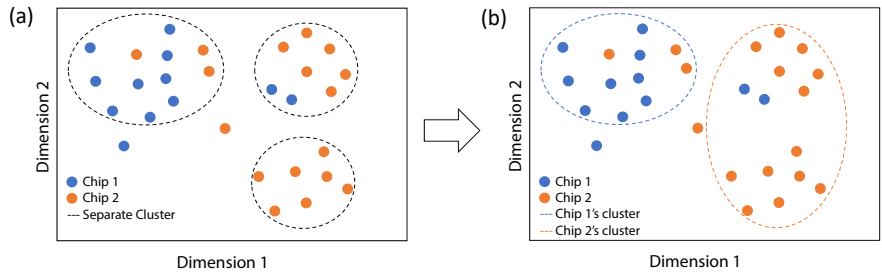

Fig. 12. Schematic of the method of combining clusters to calculate accuracy is shown for illustrative data.
(a) The clustering algorithm separates the data into three groups.
(b) For accuracy calculation, we consolidate two of the clusters as described in section 6.3.
The number of false positives in this example is 2/11 = 18% and the number of false negatives is 3/16 = 19%. Two data points (outside of dashed ellipses) are classified as noise.

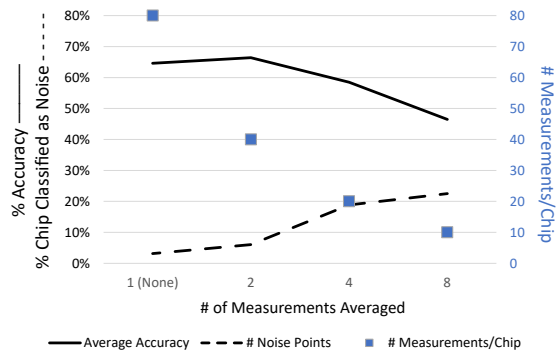

Fig. 13. Characterizing the tradeoff between various levels of averaging with PCA on the scalable trojan benchmarks.

### 6.5 Example Results

The result of applying the CNN and density based clustering to data from several test chips (Figure 10) is shown in Figure 14. Each dataset is plotted along the 2 components that allow for best visualization of that specific dataset and its clustering. However, the clustering algorithm itself uses all four dimensions of the CNN latent variable components in its calculations. Figure 14 only shows the results for the test data (33% of the total data). No trojan is detected in the case of the trojan free test chip (c). Trojans are detected in the cases of the large and small counter trojan test chips (a and b).

As discussed in the next section, the accuracy of the CNN autoencoder is higher than that of PCA.

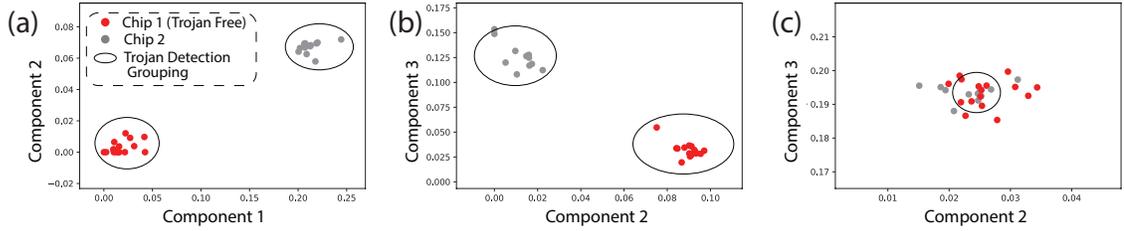

Fig. 14. Successful classification of QDM data for test chips is achieved by applying density based clustering to CNN latent variables for (a) Trojan Free circuit vs. Large Trojan test chip, (b) Trojan Free circuit vs. Small Trojan test chip, and (c) Trojan Free circuit vs. another Trojan Free test chip. Any points outside a group are a result of noise.

Table 2. PCA vs CNN Average Figures of Merit for Scalable Benchmarks
False Positives (FP), False Negatives (FN), Accuracy, and Noise.

| Method | Trojans | FP | FN | Accuracy | Noise |
| --- | --- | --- | --- | --- | --- |
| CNN | Large | 0.4/16 = 2.4% | 1.1/11 = 9.8% | 23.5/27 = **87.2%** | 7.4% |
| PCA | Large | 0/40 = 0% | 19.8/40 = 49.6% | 56.3/80 = **70.4%** | 4.8% |
| CNN | All | 0.3/16 = 1.6% | 4.1/11 = 36.8% | 20.8/27 = 76.9% | 4.8% |
| PCA | All | 3.9/40 = 9.8% | 18.2/40 = 45.4% | 53.1/80 = 66.4% | 4.1% |

## 6.6 Trojan Parameters

The scalable trojan tests evaluate the detection limits and strengths of the QDM+ML method. We consider the effect of low-power trojans by scaling each type of hardware trojan down in size and frequency, to lower the power consumption. Detection is then attempted for each of these versions of the trojan. Smaller triggers (1 or 2 bit triggers) are expected to be more difficult to detect due to the small currents. However, a practical trojan trigger would likely require more circuitry for an effective attack. The decrease in accuracy with smaller trojan sizes is seen to be the case in Table 2, where the larger trojan triggers are detected with 87% accuracy, compared to 76.9% for all trojans. If necessary, we could average over longer times to improve the SNR and detection accuracy. Table 2 also shows that on average, CNN outperforms PCA with high spatial resolution QDM measurements by over 15% on average.

The accuracies for the individual scalable trojan tests are shown in Appendix A.1. The large trojans from Table 2 are marked with an asterisk.

## 7 TRUSTHUB TROJAN DETECTION

### 7.1 Motivation

The standard test for trojan detection in the existing literature is the TrustHub benchmark set [28, 29]. Performance of the CNN detection framework using QDM imaging data is evaluated using the TrustHub trojan dataset to compare to existing methods. This also demonstrates that QDM imaging with CNN trojan identification works for trojans representative of more complex attacks. This is especially true for trojans in cryptographic circuits that are particularly harmful and tempting from an attacker's point of view.

The TrustHub Advanced Encryption Standard (AES) benchmarks with inserted trojans are selected for evaluation in this study. These contain a base circuit performing AES encryptions with trojans inserted that leak the key or cause a denial of service. Some of the trojans are always on, whereas others have a trigger that checks for a predefined state or plaintext input [28, 29].

Table 3. Descriptions and sizes of the AES TrustHub trojan benchmarks evaluated in this work.

| Benchmark | Description | Payload Primitive Count | Trigger Primitive Count | Trigger Size (%) |
| --- | --- | --- | --- | --- |
| AES-T100 | Always on, leak AES key with CDMA covert channel | 84 Reg, 9 LUT | N/A | N/A |
| AES-T300 | Always on, leak intermediate states | 64 Reg, 26 LUT | N/A | N/A |
| AES-T500 | Triggered on sequence of plaintext, reduce battery life (dos) | 128 Reg | 1 Reg, 179 LUT, 44 Carry | 1.9% |
| AES-T700 | Triggered on plaintext value, leak AES key with CDMA covert channel | 84 Reg, 9 LUT | 1 Reg, 43 LUT, 11 Carry | 0.5% |
| AES-T1400 | Triggered on plaintext sequence, leak intermediate states | 64 Reg, 32 LUT | 1 Reg, 179 LUT, 44 Carry | 1.9% |

### 7.2 FPGA Setup

The plaintext and key inputs to the AES base circuit are generated by two 128 bit LFSRs on the FPGA outside of the diamond field-of-view. These allow for consistent control of the circuit activity between the trojan-free circuit and trojan-inserted circuits. The placement and routing of the base circuit logic is kept constant between the trojan free and trojan inserted configurations to minimize differences. Constraints are also added to preserve relevant trojan logic that would be removed in synthesis or other optimizations since we are modeling an attacker adding trojans at the foundry stage. The TrustHub trojan benchmarks tested are listed in Table 3 [28, 29]. Since the payloads of three of the trojans (500, 700, and 1400) are never activated, the trigger primitive count as a percent of the AES base logic primitive count is also given.

### 7.3 QDM Measurements and Analysis

The QDM measurement setup is the same as for the scalable trojans. As before, 80 magnetic field measurements are collected for each benchmark.

The PCA and CNN autoencoder with clustering methods are applied to the TrustHub side channel QDM image data set. Since the magnitudes of current and thus magnetic field are much larger than those of the scalable trojans, no high pass filtering or averaging is necessary to increase SNR. For the larger TrustHub trojans, the entire QDM field-of-view (600 × 600 pixels) is used, and the image data is normalized due to the larger spread in magnetic field magnitudes. Due to the larger magnetic signatures, the autoencoder is simplified to just one convolutional layer and smaller fully connected layers, and the model is only trained for 1000 epochs. To allow for accurate comparison, the initial weights are randomly seeded for all the trojan test analyses.

### 7.4 Results

The CNN autoencoder is able to successfully determine that there is no trojan when considering a trojan free test chip, and also that the trojans from Table 3 are present in the trojan inserted test chips. This is true even in the case of AES-T700, where the trojan trigger (only actively switching circuitry) is not in the diamond field-of-view, since the QDM still images power delivery to the circuitry as well as the data and clock interconnect to the trigger logic.

Table 4. PCA vs CNN Average Figures of Merit for TrustHub Benchmarks.

| Method | FP | FN | Accuracy | Noise |
|---|---|---|---|---|
| CNN | 0.8/24 = 3% | 0.6/30 = 2% | 47.2/54 = 87% | 10% |
| PCA | 2.6/80 = 3% | 17.2/80 = 22% | 134.8/160 = 84% | 3% |

The average accuracies using PCA and CNN are shown in Table 4. The average accuracy does not include the AES-T700 benchmark, which has the trojan trigger outside of the diamond field-of-view. In this demonstration, the diamond location is fixed to a set field-of-view. However, a practical setup to be pursued in future work would stitch together a larger field-of-view through multiple QDM measurements across the chip or scale up the field-of-view with a larger diamond to cover the entire chip in a single image. Results for the individual benchmarks are in Appendix A.2.

## 8 DISCUSSION

Comparison between the CNN and PCA (prior work) methods in the previous sections shows that the deep learning is beneficial for trojan detection accuracy. In addition, the QDM measurement framework offers unique advantages over existing methods and provides opportunities to utilize previously inaccessible measurement regimes. High resolution and wide field-of-view imaging of the spatially distributed DC magnetic field associated with power delivery provides novel insights into the characteristics of the device under test. For example, high frequency electromagnetic fields can be shielded and suppressed due to the skin depth of common metals and spatially distributed thermal effects can be hidden through simple coatings. However, the low frequency magnetic fields detected by the QDM are challenging to shield or distort and require high permeability of ferrous materials to distort the local magnetic field and distort the characteristic magnetic fingerprints of power delivery and resource distribution.

The attacker can attempt to insert HTs such that the layouts with and without trojan are spatially similar, thus attempting to reduce the benefit of the high spatial resolution imaging of the QDM over standard near field probe methods. However, when we take into account different switching activity, component sizing, and necessary variations in metal layer routing for the inserted trojan, it is difficult to have the exact same currents and magnetic signatures in both cases, especially given the full-magnetic vector sensing capabilities of the QDM. It might be possible to imitate one component of the magnetic field through interference of local magnetic fields, but it is challenging to hide all components of the magnetic field of a HT. The greatest tool an attacker has for hiding the trojan from the QDM is increasing stand-off distance of the package and thus blurring out the characteristic features of the magnetic field. The sensitivity of the current QDM is far from the theoretical limits of performance and great strides are being made to push down the measurement noise floor. As the sensitivity of the QDM increases, there are fewer avenues to hide HTs in the noise of the measurements with drawing small currents, interfering local magnetic fields, or increasing standoff distance of the chip. Having identical magnetic fields across all of these unknown parameters while also creating a trojan that can perform something malicious is a difficult task for an attacker in the limited time at the foundry and the steady improvement of the QDM makes this an even more challenging task.

Additionally, we perform a quantitative comparison of our QDM method with standard EM field probe-based detection [15]. To limit the differences between detection results to that of just the measurement method, we also apply the same Euclidean distance golden-chip based method described in [15]. Both works perform detection of the AES-based TrustHub trojans [28], with the Euclidean distances for the TrustHub trojans, calculated according to the method outlined in [15], shown in Table 5. While there is not exact overlap in the trojan benchmarks evaluated by the two methods, many of the benchmarks are quite similar, differing only in the PRNG initialization or plaintext comparison for triggering the trojan. As described in previous sections, we consider detection of trojans where the payload is not activated during the detection process. However, [15] only attempts to detect trojans that are pre-activated, and notes that detecting un-activated trojans would require a very high detection resolution.

Table 5. Comparison of QDM (left, this work) and EM field probe (right, [15]) detection methods using Euclidean Distance

| Benchmark | Normalized Euclidean Dist. |
|---|---|
| AES | 1 |
| AES-T100 | 1.50 |
| AES-T300 | 6.41 |
| AES-T500 | 3.88 |
| AES-T700 (outside FOV) | 1.04 |
| AES-T1400 | 1.48 |

| Benchmark | Normalized Euclidean Dist. |
|---|---|
| AES | 1 |
| AES-T100 | 1.88 |
| AES-T200 | 1.69 |
| AES-T400 | 1.64 |
| AES-T700 | 2.56 |
| AES-T800 | 1.57 |
| AES-T900 | 4.81 |
| AES-T1000 | 2.88 |
| AES-T1100 | 1.86 |
| AES-T1200 | 4.54 |
| AES-T1600 | 2.64 |
| AES-T1700 | 2.93 |

We are able to achieve this higher detection resolution with the QDM, and detect un-activated TrustHub hardware trojans with the same Euclidean distance algorithm. While the variation of the distances is higher in our work, this is likely due to the larger environmental perturbations, as well as a lack of optimized spatial pre-processing in computing these distances, such as the wavelet denoising used by [15] for time-series data.

In addition, a comparison of qualitative and quantitative parameters of prior work to this study are provided in Table 6. Importantly, the QDM method is unique in its ability to provide wide field-of-view trojan detection with a golden-chip free methodology. We emphasize that this is not a perfect comparison, since the TrustHub trojan implementations are not identical across the different works, due to different synthesis tools as well as FPGA architectures (as seen in comparing the number of LUTs and Registers in our results and prior works). However, these results are a clear quantitative indication that the QDM has benefits over preexisting methods. Future work will include optimized implementations of standard EM probe test setups for direct comparison. This will allow for side by side comparison of the measurement capabilities.

In summary, we find that the QDM + CNN framework can detect TrustHub trojan triggers effectively, with a minimum size of 0.5% of the total logic primitive count as defined in Table 3. We expect the sensitivity of the QDM will improve further with ongoing experimental improvements, allowing for detection of even smaller trojans in the future.

## 9 CONCLUSION

In this work, we demonstrated the first usage of high spatial resolution and wide field-of-view magnetic field measurements using the Quantum Diamond Microscope (QDM) for hardware trojan detection. We also showed that this side channel data can be used for trojan detection through a convolutional neural network (CNN) and clustering analysis for unsupervised deep learning.

Table 6. Comparison to Prior work. (NF = Near Field, BM = Benchmark, EU = Euclidean Distance)

| Metric | This Work | [35] | [6] | [24] | [15] | [2] | [11] |
|---|---|---|---|---|---|---|---|
| **Type** | Side Channel | Side Channel | Side Channel | Side Channel | Side Channel | Side Channel | Post RTL |
| **Data Collection** | QDM | Thermal | Power | Backscattering | NF Probe | NF Probe | Netlist |
| *Golden Reference Needed* | No | Yes | No | No | Simulated Spectra | Yes | Yes |
| **Analysis** | CNN + Cluster | INN | Cluster | Cluster | EU | T-Test | PCA |
| **DUT** | Artix-7 | 130nm Sim | Kintex-7 | Cyclone V | Spartan-6 | Virtex-II | N/A |
| *Instantaneous FOV & Resolution* | 3.7 × 3.7 mm 600×600 pixels | Full Chip 20×20 pixels | Single Point | Single Point | Single Point | Single Point 17×21 grid | N/A |
| **Test Logic** | TrustHub AES BM | AES w/ sequential HT | TrustHub AES BM | TrustHub AES BM | TrustHub AES BM | AES-128 w/ comparator HT | TrustHub BM |
| **Result** | 87% Accuracy 3% FP 2% FN | 98.2% Accuracy | 88.75% Accuracy | 100% Accuracy | EU > Threshold for all HTs | Visual comparison shows HTs | 96% Accuracy 3% FP 58% FN |

An initial result of these measurements is that there is more DC current leakage of information from integrated circuits (ICs) than initially expected, as represented by patterns of magnetic fields induced by the currents, which pass out of ICs largely undisturbed. This DC leakage would be difficult to see in global power measurements due to other ongoing activity and is not observable in inductive techniques, like EM probes, because the induced signal is small and vanishing at low frequencies .

The present work on FPGA based hardware trojans can easily be extended to custom integrated circuits that have a similar fabricated structure. One major difference is that FPGAs can be reprogrammed to express both trojan free circuits and many trojan inserted circuits with a single chip. However, many circuits of interest are ASICs, where a trojan free circuit will be a different physical component from a trojan inserted circuit.

In addition, it is important to test the efficacy of the QDM+ML trojan detection method with respect to process variation, for example with multiple FPGA test chips. This will be addressed in a future study. Future improvements of the QDM setup include enhanced magnetic field sensitivity, faster measurement speed, and increased field-of-view. Measurements are currently limited by the photoelectron-per-second capacity of the cameras, which scales as pixel well depth × number of pixels × sampling rate. Improved cameras such as lock-in cameras [32], brighter diamonds, and higher collection efficiency will further improve measurement performance.

## ACKNOWLEDGMENTS

This project was fully funded by the MITRE Corporation through the MITRE Innovation Program. M.A. acknowledges support from the NSF GRFP (Grant No. 1745302), Analog Devices Fellowship, and the MathWorks Engineering Fellowship. M.J.T. and R.L.W. acknowledge support from the Quantum Technology Center (QTC) at the University of Maryland. NV-diamond sensitivity optimization pertinent to this work was partially supported by the DARPA DRINQS program (Grant No. D18AC00033)

Table 7. False positive (FP) and false negative (FN) values for scalable trojan detection using PCA and CNN analyses of QDM magnetic imaging data.

| Trojan Free Circuit | Trojan in Test Chip | PCA FP | PCA FN | CNN FP | CNN FN |
|---|---|---|---|---|---|
| 200 bit counter | None* | 0% | N/A | 0% | N/A |
| 167 bit LFSR | None* | 0% | N/A | 0% | N/A |
| 200 bit counter | 8 bit counter* | 0% | 0% | 0% | 0% |
| 200 bit counter | 4 bit counter* | 0% | 100% | 0% | 0% |
| 200 bit counter | 2 bit counter | 55% | 0% | 0% | 100% |
| 200 bit counter | 1 bit counter | 0% | 63% | 0% | 64% |
| 200 bit counter | 8 bit shift register* | 0% | 0% | 0% | 0% |
| 200 bit counter | 4 bit shift register* | 0% | 0% | 0% | 0% |
| 200 bit counter | 2 bit shift register | 70% | 0% | 0% | 91% |
| 200 bit counter | 1 bit shift register | 0% | 55% | 0% | 100% |
| 200 bit counter | 32 bit comparator | 70% | 0% | 0% | 64% |
| 200 bit counter | 16 bit comparator | 0% | 65% | 0% | 100% |
| 200 bit counter | 8 bit counter (1/2 frequency)* | 0% | 0% | 0% | 9% |
| 200 bit counter | 8 bit shift register (1/2 frequency)* | 0% | 60% | 0% | 0% |
| 200 bit counter | 8 bit shift register (1/4 frequency) | 0% | 80% | 0% | 91% |
| 167 bit LFSR | 8 bit counter* | 0% | 100% | 0% | 0% |
| 167 bit LFSR | 4 bit counter* | 0% | 100% | 0% | 0% |
| 167 bit LFSR | 4 bit shift register* | 0% | 85% | 0% | 100% |
| 167 bit LFSR | 80 bit comparator* | 0% | 100% | 0% | 9% |
| 167 bit LFSR | 32 bit comparator* | 0% | 100% | 31% | 9% |

Table 8. False positive and false negative values for TrustHub trojan detection using PCA and CNN analyses of QDM magnetic imaging data.

| Trojan Free Circuit | Trojan in Test Chip | PCA FP | PCA FN | CNN FP | CNN FN |
|---|---|---|---|---|---|
| AES | None | 16% | N/A | 0% | N/A |
| AES | AES-T100 | 0% | 0% | 8% | 3% |
| AES | AES-T300 | 0% | 8% | 0% | 7% |
| AES | AES-T500 | 0% | 100% | 0% | 0% |
| AES | AES-T700 (outside FOV) | 0% | 99% | 13% | 27% |
| AES | AES-T1400 | 0% | 0% | 8% | 0% |

## A INDIVIDUAL TROJAN ACCURACIES

### A.1 Scalable Trojans

Table 7 shows PCA and CNN performance on QDM data sets in terms of false positives and false negatives for each individual scalable trojan test from Section 6.6.

### A.2 TrustHub Trojans

Table 8 shows PCA and CNN performance on QDM data sets in terms of false positives and false negatives for each individual TrustHub trojan test from Section 7.4.

## B NEURAL NETWORK ARCHITECTURES

### B.1 Scalable Trojans

The CNN layer architecture used in Section 5.3 to detect scalable trojans is shown in Figure 15.

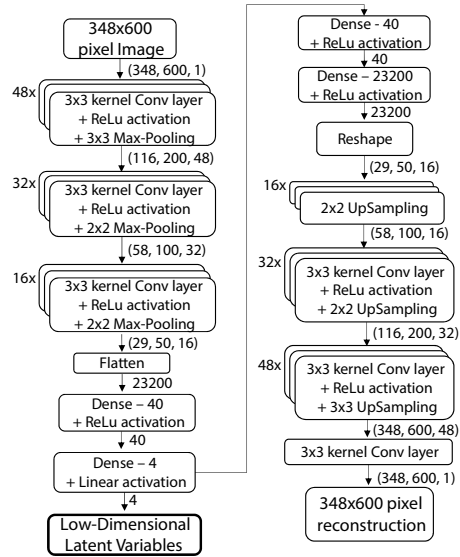

Fig. 15. Block diagram of the CNN autoencoder architecture used for scalable trojan detection. Training is done to minimize error between the original image and the reconstruction. The low dimensional features are used as inputs to the clustering algorithm for trojan detection. The Max-Pooling layers are used to reduce spatial dimensions and avoid overfitting.

### B.2 TrustHub Trojans

The CNN layer architecture used in Section 7.3 to detect TrustHub trojans is shown in Figure 16.

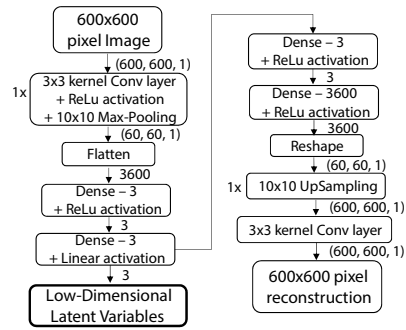

Fig. 16. Block diagram of the simplified CNN autoencoder architecture for TrustHub trojan detection.

## C QUANTUM DIAMOND MICROSCOPE EXPERIMENTAL SETUP (FIGURE 17)

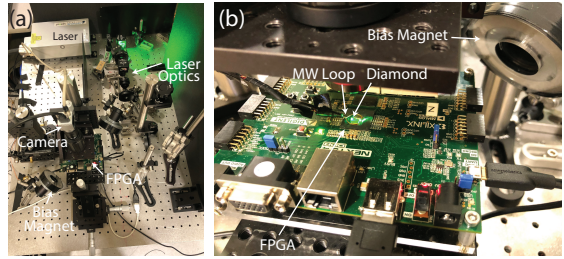

Fig. 17. (a) Complete optical setup of QDM; (b) Development board with diamond placed over field-of-view in decapsulated FPGA. Diamond, microwave loop, laser, and camera positioned above the FPGA for ODMR measurements.

## D TROJAN PSEUDOCODE FOR COMPARATOR, SHIFT REGISTER, AND COUNTER BUILDING BLOCKS

### D.1 Comparator Trojan Building Block

**Require:** in (4 bits from base code), enable
    trigger ← ((enable == '1') AND (in == '1111'))

### D.2 Shift Register Trojan Building Block

**Require:** in (1 bit from base code), clock, reset, enable
    **if** rising edge clock **then**
        **if** reset **or not** enable **then**
            shift_reg ← 0
        **end if**
        shift_reg[3:1] ← shift_reg[2:0]
        shift_reg[0] ← in
    **end if**
    trigger ← (shift_reg == '1111')

### D.3 Counter Trojan Building Block

**Require:** in (1 bit from base code), clock, reset, enable
    **if** rising edge clock **then**
        **if** reset **or not** enable **then**
            count ← 0
        **end if**
        **if** in = '1' **then**
            count ← count + 1
        **end if**
    **end if**
    trigger ← (count == '1111')